\documentclass{article}

\PassOptionsToPackage{numbers}{natbib}

\usepackage[preprint]{neurips_2022}

\usepackage[utf8]{inputenc} 
\usepackage[T1]{fontenc}    
\usepackage{hyperref}       
\usepackage{url}            
\usepackage{booktabs}       
\usepackage{amsfonts}       
\usepackage{nicefrac}       
\usepackage{microtype}      
\usepackage{xcolor}         

\usepackage{graphicx}
\usepackage{caption}
\usepackage{subcaption}
\usepackage{siunitx,booktabs}
\usepackage{booktabs} 
\usepackage{amsmath}
\usepackage{bbm}
\usepackage{multirow}

\usepackage{amsmath}
\usepackage{bbm}
\usepackage{multirow}
\usepackage{array}

\usepackage{amsmath}
\usepackage{amssymb}
\usepackage{mathtools}
\usepackage{amsthm}

\usepackage[capitalize,noabbrev]{cleveref}

\theoremstyle{plain}

\theoremstyle{definition}

\theoremstyle{remark}

\usepackage{algorithm,algorithmic}
\usepackage{hyperref}

\newcommand{\sq}[1]{#1}
\newcommand{\zheng}[1]{#1}

\title{Error-aware Quantization through Noise Tempering} 

\author{Zheng Wang$^*$, Juncheng B Li\thanks{\ \ Co-first author}\ ,       Shuhui Qu, Florian Metze, Emma Strubell\\
\{zhengwan, junchenl, fmetze, estrubel\}@cs.cmu.edu}

\begin{document}

\maketitle

\begin{abstract}
Quantization has become a predominant approach for model compression, enabling deployment of large models trained on GPUs onto smaller form-factor devices for inference.
Quantization-aware training (QAT) optimizes model parameters with respect to the end task while simulating quantization error, leading to better performance than post-training quantization. Approximation of gradients through the non-differentiable quantization operator is typically achieved using the straight-through estimator (STE) or additive noise.
However, STE-based methods suffer from instability due to biased gradients, whereas existing noise-based methods cannot reduce the resulting variance. 
In this work, we incorporate exponentially decaying quantization-error-aware noise together with a learnable scale of task loss gradient to approximate the effect of a quantization operator. We show this method combines gradient scale and quantization noise in a better optimized way, providing finer-grained estimation of gradients at each weight and activation layer's quantizer bin size. 
Our controlled noise also contains an implicit curvature term that could encourage flatter minima, which we show is indeed the case in our experiments.
Experiments training ResNet architectures on the CIFAR-10, CIFAR-100 and ImageNet benchmarks show that our method obtains state-of-the-art top-1 classification accuracy for uniform (non mixed-precision) quantization, out-performing previous methods by 0.5-1.2\% absolute.
\end{abstract}

\section{Introduction}
\label{sec:intro}
Driven by advantages in scalability, privacy, low latency and cost, machine learning ``on the edge'' is garnering increased interest and application in diverse areas.
However, modern state-of-the-art (SOTA) machine learning models based on dense neural networks are too large to run on edge devices, where memory and compute cycles are limited compared to the servers where models are typically trained and deployed~\cite{gholami2021survey}.
Quantization has emerged as an effective approach to compress full (32-bit) precision models by reducing the number of bits used to represent each model parameter. Unlike other methods that boost efficiency of ML models, quantization does not require altering the original model architecture or pruning weights~\cite{neill2020overview}~\cite{gholami2021survey}.

\begin{figure}[t]
    \centering
    \includegraphics[width=1.0\linewidth]{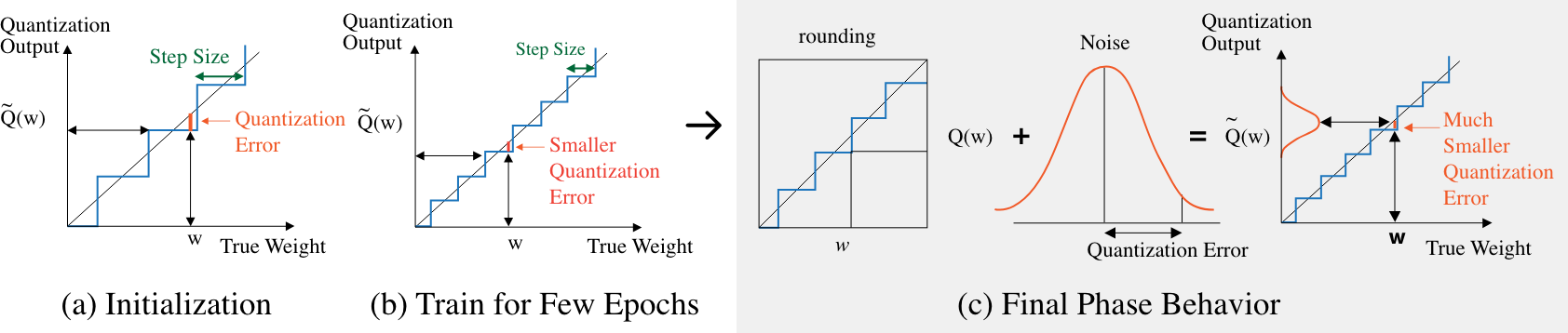}
    \caption{An overview of our method: (a) Models are initialized with large quantization bin widths and correspondingly high quantization error. (b) After a few epochs the bin widths will reduce, increasing rounding accuracy and decreasing quantization error. (c) When quantization error becomes small enough, quantization noise kicks in to fine-tune model weights.}
    \label{fig:overall}
\end{figure}

Typically, the full-precision weights and/or activations of the neural network model are discretized by a quantization function. 
Post-training quantization can be performed on any existing model, but at the severe cost of accuracy, particularly at lower precisions such as 4 or 2 bits.  Quantization-aware training mitigates this loss in accuracy by simulating quantization of full-precision weights and activations during training, allowing for model parameters to be optimized such that they form accurate predictions even when quantized. 
Quantization approaches can be further categorized into uniform and non-uniform quantization. Non-uniform quantization approaches have been shown to obtain better accuracy than uniform quantization~\cite{yamamoto2021learnable}\cite{liu2021nonuniformtouniform}, but they are not amenable to deployment on existing hardware, which would require special support for the specific mixed-precision schemes (e.g. additional data structures to store codebooks or quantization intervals \cite{gholami2021survey}). In this work we therefore focus on the more practical setting of \emph{uniform} quantization via quantization-aware training (QAT). 

The key challenge in QAT is approximating gradients to the non-differentiable quantization function. 
QAT methods typically quantize all the weights during the forward pass and introduce the straight through estimator~(STE; \cite{bengio2013estimating}) to estimate the gradient of these non-differentiable layers. To be concrete, let $\mathbf{Q}(w)$ be the non-differentiable quantization function and $Q_L$ and $Q_H$ be the lower and upper bound in an $n$-bit system. STE sets the gradient $\tfrac{\partial \mathbf{Q}(w)}{\partial w} = 1$ for all $Q_L \le w \le Q_H$ and 0 otherwise. The gradient estimate of loss function $\mathcal{L}$ with respect to weight $w$ is:
\begin{equation}
\left. \frac{\partial \mathcal{L}}{\partial w}\right|_{w} \approx
\begin{cases}
\left. \frac{\partial \mathcal{L}}{\partial \mathbf{Q}(w)}\right|_{\mathbf{Q}(w)} & \text{if}\ Q_L \le w \le Q_H\\
0 & \text{otherwise}
\end{cases}
\label{eqn:ste_general}
\end{equation}
This approach works reasonably well when the errors introduced by STE are small \cite{yin2019understanding} e.g. quantizing 32-bit representations to 16 or 8 bits. However, since the gradient estimation shown in Eq.~(\ref{eqn:ste_general}) does not exactly reflect the true gradient, it introduces bias and instability to training. 
Moreover, due to the deterministic property of such gradient estimation, there can be accumulation of periodic quantization error, as is well studied in the field of signal processing~\cite{smith1983comparison, taheri2008}.
Specifically, as \cite{defossez2021differentiable} illustrated,
$\mathbf{Q}(w)$ oscillates between the quantized value just above $(w_{+})$ and just under $(w_{-})$ the unquantized ground truth, $w_{*}$, while $w$ oscillates around the boundary $(w_{+} + w_{-})/2$.

One method that has been proposed to address these challenges is by introducing additional parameters allowing for finer grained estimates. Learned step size quantization~(LSQ; \cite{esser2019learned}) has been proposed to address these challenges by learning the quantization bin width, or \emph{step size}, as a model parameter. 
The learned step size naturally provides more precision in optimization, however LSQ still falls within the regime of STE and thus still ultimately suffers from the same bias and instability issues, albeit to a lesser extent.
An alternative approach to estimate the gradient during quantization-aware training is to introduce noise that simulates the quantization process.
\cite{defossez2021differentiable, sakr2018per} introduce a psuedo-noise quantizer, $\tilde{\mathbf{Q}}(w) = w + noise$, inspired by analog-to-digital converter (ADC) simulation. 
The key advantage of such an additive noise approach is that it provides a potentially unbiased estimator of the true gradient when there is limited knowledge of the distribution of $\mathbf{Q}(w)$. Under the local linearity assumption:
$
\mathbb{E}\left( \left. \frac{\partial \mathcal{L}}{\partial \tilde{\mathbf{Q}}(w)}\right|_{w + noise}\right) = \left. \frac{\partial \mathcal{L}}{\partial \tilde{\mathbf{Q}}(w)}\right|_{\mathbb{E}[w + noise]} = \left. \frac{\partial \mathcal{L}}{\partial w} \right|_{w}
$
~\cite{baskin2021nice} showed such noise may improve convergence. However, because the linearity assumption does not hold when the noise is large, and because simulating the quantization error requires the variance of noise to increase with model compression, the gradient estimator in fact does not reflect true gradient throughout the compression process.

To propose a remedy, we revisit the theoretical foundation of SGD, ~\cite{bayesSGD2017} proposed a framework formulating the gradient update as $W_{t+1} = W_{t} - \eta(\nabla \mathcal{L}(W_{t}) + \varepsilon(t))$, $\varepsilon(t)$ denotes random noise from Gaussian family updating at step $t$ towards convergence, which models the effect of estimating the gradient using mini-batches.
They also proved that the "temperature" (which can be treated as the magnitude of the variance of mini-batch gradient), $T$, is approximately $\eta / B$, where $\eta$ is the learning rate and $B$ the batch size. 
Intuitively, the initial noisy phase allows the model to explore a larger fraction of the parameter space without getting trapped in sharp local minima. 
Once we find a promising region of parameter space, we reduce the noise to fine-tune the parameters through LR decay. However, this formulation is for vanilla SGD without quantization.

Following such intuition, in this work, we design an additive noise to interact with quantization error during the different stages of SGD weight updates.
We propose combining learned step size with additive noise using carefully controlled variances to further improve the performance of quantized models.
Early during training when mini-batch gradient has large variance, we train using STE with variable step size.
In the later stages of training once variances are small enough, quantization error aware noise kicks in to provide finer-grained updates. 
We will show such noise also tends to encourage the ``flatness" of the trained model, since it contains a second-order curvature term. 
Figure~\ref{fig:overall} illustrates the overall process of our approach. 
Our experiments on ResNet architectures trained on the CIFAR-10, CIFAR-100, and ImageNet datasets demonstrate better accuracy than previous state-of-the-art uniform quantization approaches.


Our contributions are listed as follows:
\begin{enumerate}
    \item We propose a novel algorithm for quantization-aware training by combining gradient scale and quantization noise, which improves state-of-the-art uniform quantized model performance without introducing additional parameters or observable computation cost.
    \item We provide extensive experimental analysis of our approach, motivated by a detailed theoretical analysis of the quantization and optimization process. 
    \item We show that quantized model trained with our method tend to have flatter minima, which is favorable for generalization to unseen data.

\end{enumerate}

\section{Related Work}
\label{related}
Pioneering work in quantization~\cite{ courbariaux2015binaryconnect, zhou2016dorefa} looked at quantizing model weights, activations, and gradients to accelerate neural network training. A recent surge of interest in quantization research is driven by the need to deploy ML models onto edge devices;  \cite{neill2020overview} and \cite{,gholami2021survey} provide comprehensive overviews of the field to date. 
Recently, there is also an renewed interest in post-training quantization (PTQ). \cite{nagel2020up} analyzed the loss degradation by
Taylor expansion, which inspired \cite{li2021brecq} to formulate a second-order approximation. \cite{cai2020zeroq} leveraged synthetic data to fine-tune the pre-trained model, and \cite{nahshan2020loss} analyzed into the loss landscape. However, PTQ methods still lag behind quantization-aware training (QAT) methods in accuracy, which are the focus of this work.
The terminology QAT was first introduced by~\cite{jacob2018quantization}. QAT incoporates quantization error as 
part of the overall loss minimized during optimization. 

\subsection{Learnable quantization parameters}
Several works proposed learning-based approaches to improve QAT around the same time: \cite{jain2019trained} proposed to learn the quantizers’s dynamic range while training; \cite{uhlich2019mixed} advocated for learning the optimal bit-width. 
Among them, LSQ~\cite{esser2019learned} and its extension LSQ+ \cite{bhalgat2020lsq} were the simplest and best performing approaches, which propose learning the optimal quantizer step size (bin width).
From there, the community \cite{stock2020and} has focused on minimizing the quantization error rather than merely preserving the weights.
These methods have all observed only minor performance drops compared to full-precision models, but nevertheless inherit the same instability and 
bias issues of STE~\cite{fan2021training}.
In this work, we propose to alleviate this issue by carefully mixing STE with noise throughout the training process.

\subsection{Quantization noise}
Additive noise to quantization in neural network models was first studied by \cite{baskin2021nice,sakr2018per}, and \cite{fan2021training} crafted dropout-based noise by sampling at each layer and training step whether to use
the quantized or unquantized weights. \cite{defossez2021differentiable} leveraged ADC noise to completely replace the usage of STE.
Additive noise based approaches avoid the risk of systematic bias but introduces variances to the system~\cite{sakr2018per}.
There is a strong motivation for us to leverage noise to reduce the bias from STE, while controlling the variance of the noise itself.

\textbf{Uniform quantization versus non-uniform quantization}
The above-mentioned techniques are all \emph{uniform quantization}, which builds a mapping from real values to integer values, resulting in uniformly spaced quantization values. Naturally, the non-uniformly spaced counterpart is called \emph{non-uniform quantization}, and involves dequantization step. 
Recent works \citep{yamamoto2021learnable, liu2021nonuniformtouniform} showed competitive performance, since they could better capture the quantized number
distributions and focus more on important value regions.
But they all require extra storage and tweaking the data structure, which limited the methods practicality to be deployed, given the exisiting DL software and hardware paradigm.~\cite{gholami2021survey}

\section{Methodology}
\subsection{Stochastic Gradient Descent with momentum + Quantization}
\label{sec:SGD}
It is known that model trained with SGD tends to generalize better than those trained with full-batch gradient descent. Recent works \cite{dontdecaylr2017, bayesSGD2017, noiseSGD2020, randomnessSGD2021, regSGD} theoretically explain this phenomenon, showing that, as a consequence of Central Limit Theorem, the mini-batch gradient used in SGD, $\nabla \hat{\mathcal{L}}(W_t)$, can be treated as the clean full-batch gradient, $\nabla \mathcal{L}(W_t)$ plus an additive Gaussian noise term evolving over time $\varepsilon(t)$ (briefly introduced in \S~\ref{sec:intro}) with mean of 0 and variance approximately of $\Sigma(W_t)/B$. Here, the term $\Sigma(W_t)$ is the graident covariance matrix, which is a function of current parameter values. Thus, we can express the SGD update with the form $W_{t+1} = W_{t} - \eta(\nabla \mathcal{L}(W_{t}) + \varepsilon(t))$, where $W_{t}$ is the collection of trainable parameters at step $t$, $\eta$ the learning rate, $B$ is the batch size, and $\mathcal{L}(\cdot)$ the loss function. Then suppose we define "temperature" $T = \eta/B$, we can then draw the noise term, $\varepsilon_\Phi$ from standard normal distribution $\mathcal{N}(0,I)$, and arrive at the following equivalent form:
\begin{equation}
    W_{t+1} = W_{t} - \eta \nabla \mathcal{L}(W_{t}) + \sqrt{\eta T} \cdot \Sigma(W_t)^{1/2} \cdot \varepsilon_\Phi
\end{equation}

Here the temperature term $T$ controls the magnitude of the noise in SGD graident and it is proven by \cite{randomnessSGD2021, bayesSGD2017} that a higher temperature promotes the model to converge to a flatter minima in the loss landscape. This is important in QAT processes. While in quantized models, the SGD update follows a similar pattern, because the quantization function $\mathbf{Q}(W)$ is usually non-differentiable, a common practice is to use straight through estimator \cite{bengio2013estimating} to pass the gradient from $\mathbf{Q}(W)$ to $W$, as described in Eq. \ref{eqn:ste_general}. Assume that $W' \subseteq W$ such that $Q_L  \le W' \le Q_H$ (so that the STE gradient is 1, the elements in $W \setminus W'$ has STE gradient of 0, which are not updated), then the SGD update in a quantized model is:
\begin{equation}
    \begin{aligned}
        W_{t+1}' &= W_t' - \eta \frac{\partial \hat{\mathcal{L}}}{\partial \mathbf{Q}(W)} \frac{\partial \mathbf{Q}(W)}{\partial W} = W_t' - \eta \left(\nabla_{\mathbf{Q(W)}} \mathcal{L} + \varepsilon(t) \right) \\
        &= W_t' - \eta \nabla_{\mathbf{Q}(W)} \mathcal{L} + \sqrt{\eta T}\cdot \Sigma(W_t')^{1/2} \varepsilon_\Phi
    \end{aligned}
    \label{sgd_quant}
\end{equation}
\looseness=-1Here, we take the $\Sigma(W_t')$ as the gradient covariance matrix after marginalizing out $W\setminus W'$ in the original noise term $\varepsilon$. The major difference in the update comparing to full precision model is the replacement of the clean full-batch gradient term $\nabla_W \mathcal{L}$ with $\nabla_{\mathbf{Q}(W)} \mathcal{L}$, which results in biased gradient estimation.

Thus, there is a stronger need for such flatness for those parameters which incur high quantization error, namely when $|\mathbf{Q}(w)-w|$ is large. This is because larger quantization error leads to less accurate update in Eq. \ref{sgd_quant}, and consequently, the parameter $w$ can get stuck at some sub-optimal value near the true minima, $w^*$, as mentioned in \cite{defossez2021differentiable}. One way to alleviate the negative impact of this behavior is to prevent the model from getting stuck at sharp minima, where the loss incurred due to small deviation can be more significant. In order to achieve this, we introduce an extra quantization-error-dependent temperature term $T_Q$ ($T,\ T_Q \propto \text{Var}[\varepsilon(t)]$ and $T_Q$ and $T$ has same order) as the following:
\begin{equation}
    W_{t+1}' = W_t' - \eta \nabla_{\mathbf{Q}(W)} \mathcal{L} + \sqrt{\eta T}\cdot \Sigma(W_t')^{1/2} \varepsilon_\Phi + \eta \gamma_Q\sqrt{T_Q} \tilde{\varepsilon}_\Phi
\label{q_temp}
\end{equation}
\looseness=-1 where $\varepsilon_\Phi$ and $\tilde{\varepsilon}_\Phi$ are independent standard Gaussian noise sources, and $\gamma_Q$ is a scaling function of the quantization-error-dependent noise term. This allows the model to dynamically adjust the gradient temperature during the QAT process and encourages the model to choose flatter minima, which are more robust even if parameter update get stuck at sub-optimal values.

In case of SGD with momentum, the momentum accumulation will introduce a scaling factor $\alpha(m)$ (a function of the momentum coefficient $m$) to the noise term. In Appendix A.1 we provide a derivation showing that $\alpha(m) \approx \frac{1}{1+m^2}$. Defining the true accumulated gradient of SGD with momentum as $V_T = \sum_{t=0}^{T} m^t \cdot\nabla_{\mathbf{Q}(W)}\mathcal{L}$, the update in SGD with momentum is 
\begin{equation}
    W_{t+1}' = W_t' - \eta V_t + \sqrt{\frac{\eta T}{1-m^2}}\cdot \Sigma(W_t')^{1/2} \varepsilon_\Phi + \eta \sqrt{\frac{T_Q}{1-m^2}}\gamma_Q \tilde{\varepsilon}_\Phi
\label{sgd_m}
\end{equation}

\subsection{Quantization function}
In order to use the idea of SGD tempering to improve the quantization convergence, we solve a particular case of Eq. \ref{sgd_m} with the following setting of scaling function $\gamma_Q$ and quantization-error-dependent temperature $T_Q$:
\begin{equation}
\begin{aligned}
    \gamma_Q &= c \cdot \exp \left(-k|\mathbf{Q}(W)-W|\right) \nabla^2_{\mathbf{Q}(W)}\mathcal{L}\\
    T_Q &= |\mathbf{Q}(W) - W|
\end{aligned}
\label{coefficient}
\end{equation}
where $0<c<1$ and $k>0$ are hyperparameters. For the temperature term, we use the simplest quantization-error-based function. i.e. the magnitude of the quantization error. $|\mathbf{Q}(W) - W|$. 

The scaling function seems to be more complex when expanded out in gradient noise. However, it allows us to write a clean quantization function as in Eq. \ref{eqn:ours} that approximately give this desired gradient noise. We provide a proof to this in Appendix A.2. Furthermore, the second order derivative term $\nabla^2_{\mathbf{Q}(W)}\mathcal{L}$ provides a powerful mechanism to lower the quantization error aware tempering term even when the quantization error is large. This happens when the model already ends up close to a flat minima, corresponding to smaller second order derivative term. In such case, as the model approaches the desired local optimal in the loss landscape, it is desirable to reduce the temperature of the gradient noise to allow better convergence.
\begin{align}
    \tilde{\mathbf{Q}}(W) = \mathbf{Q}(W) + \text{sg} \left( c \cdot \exp({-k| \mathbf{Q}(W) -W |}) \cdot \sqrt{| \mathbf{Q}(W) -W |}  \cdot  \varepsilon_\Phi  \right)
    \label{eqn:ours}
\end{align}

In order to test the effectiveness of quantization error aware tempering, we choose $\mathbf{Q}(W)$ in Eq. \ref{eqn:ours} as the step-size-based quantizer in Eq. \ref{sq}, which is an effectively yet simply quantizing technique commonly adapted \cite{pact, qil2018, esser2019learned}. We also use the stop gradient operator $\text{sg}(\cdot)$ to prevent any gradient from the noise term to back-prop back to the weights.
\begin{equation}
    \mathbf{Q}(W) = \lfloor \text{clip}(W/s, Q_L, Q_H) \rceil \cdot s
    \label{sq}
\end{equation}

Here, the $Q_L$ and $Q_H$ are the lowest and highest integer in $N$-bit setting. We let the quantization step-size, $s$ to be trainable in our model and assign one step-size parameter per module. The gradient of $s$ are computed using the STE and thus has the following element-wise derivative as in \cite{esser2019learned}. Our noise $\varepsilon_\Phi$ is sampled from an isotropic Guassian distribution, i.e. $\varepsilon_\Phi \sim \mathcal{N}(0, I)$.
\begin{equation}
    \frac{\partial \tilde{\mathbf{Q}}(w_i)}{\partial s}=\begin{cases}
        -w_i/s + \lfloor w_i/s \rceil & \text{if $-Q_L < w_i < Q_H$}\\
        -Q_L & \text{if $w_i \le -Q_L$}\\
        Q_H & \text{if $w_i \ge Q_H$}
    \end{cases}
    \label{update}
\end{equation}
\vspace{-0.2cm}
\begin{figure}
    \centering
        \centering
        \includegraphics[width=\textwidth]{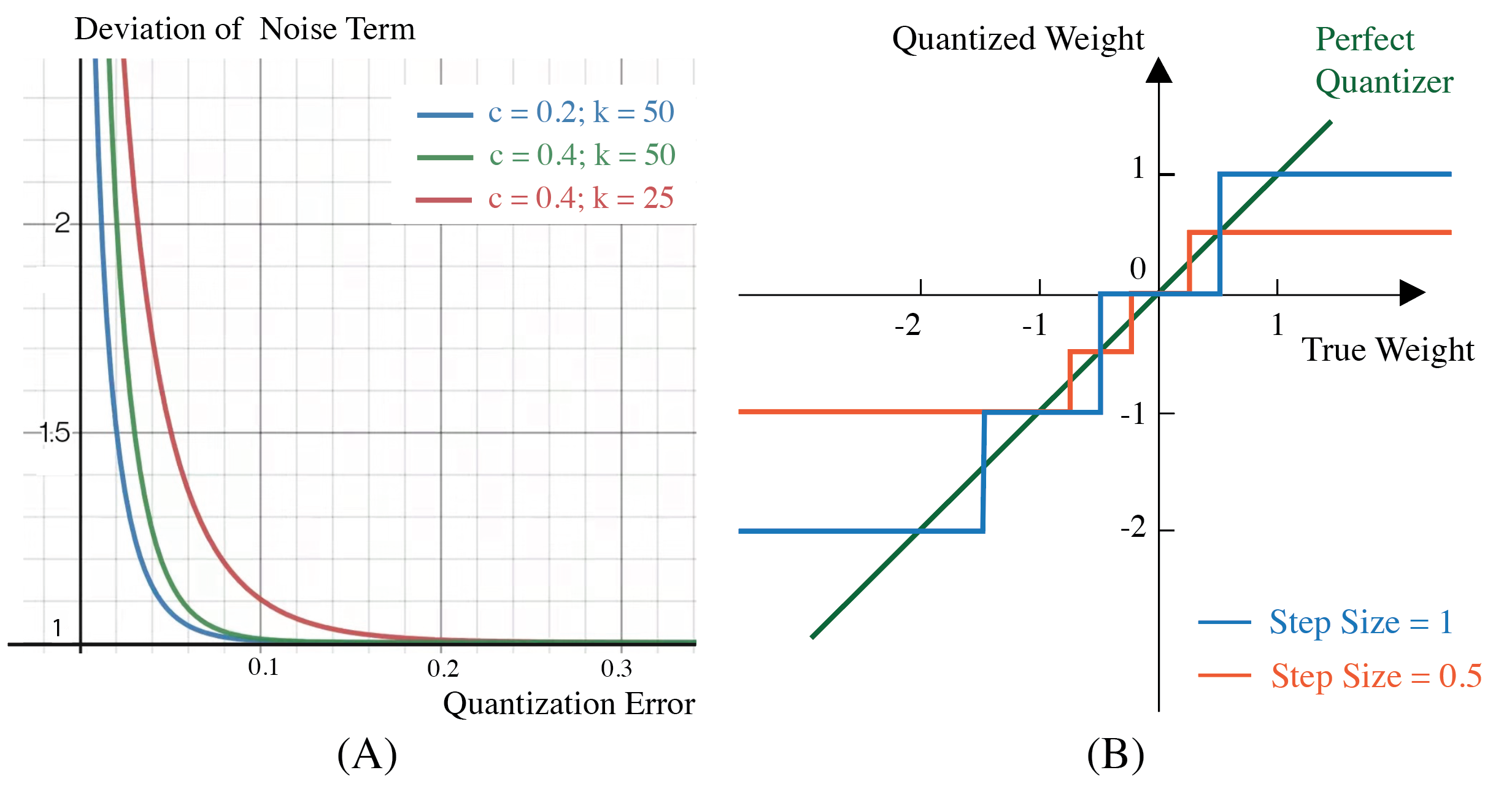}
        \label{fig:noiseVSScale}
    \hfill
    \caption{(A) Deviation of noise term versus quantization error. This figure shows the noise is trivial only when quantization error become smaller, which tends to happen when step-size parameter become lower. (B) The effect of different step sizes in quantization, the flat lines indicate value clamping. As the step size become smaller, the model can trade weight ranges for more accuracy in the gradient estimation, i.e. closer to the perfect quantizer, but get clipped with smaller value. At the same time, the quantization error will also decrease.}
    \label{step-size_noise}
\end{figure}

\subsection{Model Training Progress}
\paragraph{Initial Phase}
During the initial training phase, the quantization errors are usually fairly large, \sq{i.e. $\mathbf{Q}(w)-w\vert$ is relatively large with a high probability} and our quantization method would limited effect in noise temperature as shown in Figure \ref{fig:overall} (a). Since the hyper-parameter exponential scaling factor $k$ should be set as a relatively large integer, \sq{such as 50 in our later experiments}, we control the additive noise to be a small value through scaling with the exponential term in equation~(\ref{eqn:ours}).

In practice, with a large scaling, such as $k=50$, the additive noise term becomes significant only when $|\mathbf{Q}(w)-w| \gtrapprox 10^{-2}$. This phenomenon is illustrated in Figure~\ref{step-size_noise} (A), the graph is plotted by $y = \frac{x+c \cdot \exp{(-kx)\sqrt{x}}}{x}$ where $x=|\mathbf{Q}(w)-w|$ is the quantization error. ~\zheng{To illustrate effect of noise we set it at 1 standard deviation from the mean.}

\paragraph{Deeper Into Finetuning} 
As the QAT training progresses, the quantization step-size tends to decrease together with  the error in quantization. However, due to the necessity of maintaining large enough weights, the step-size cannot goes infinitely close to 0 since then all weights will be clipped as shown in Figure \ref{step-size_noise} (B). Depending on the choice of hyperparameter $k$, when the step-size $s$ becomes small enough, the exponential term in the scaling factor $\exp({-k| \mathbf{Q}(W) -W |})$ will be close to 1, adding the quantization-error-aware temperature to the gradient noise will encourage the model to find flatter basin in the loss landscape. Lastly, to help the final convergence of the model, we allow the learning rate as a factor to scale quantization-error-aware temperature and use learning rate schedule to stabilize the model at termination phase of training.
\section{Experiments \& Results}
We perform experiments on CIFAR-10, CIFAR-100, and ImageNet dataset to verify the effectiveness of our proposed quantization.
The CIFAR-10/100 dataset contains 50k training images and 10k test images, with 10/100 classes.
The ImageNet dataset contains 1.2M training images and 50k test images, with 1,000 classes.
We compared our results with the state-of-the-art uniform methods using several classical Computer Vision (CV) models, including ResNet-18, ResNet-34, ResNet-50, and WideResNet. We listed our experiment setup details in Appendix A.3.

\begin{table}[t]
\caption{Results on CIFAR datasets. \textbf{FP} indicates accuracy in the full precision case, $*$ indicates the model has variable bit-width and thus score with equivalent model size under that precision level is reported, and --- indicates no reported result.}
\centering
\begin{tabular}{c|l|*{3}{w{c}{1.2cm}}|*{3}{w{c}{1.2cm}}}
\hline
\multirow{2}{*}{Model}      & \multicolumn{1}{c|}{\multirow{2}{*}{Method}} & \multicolumn{3}{l|}{CIFAR10 Acc. @ Precision} & \multicolumn{3}{l}{CIFAR100 Acc. @ Precision} \\
                            & \multicolumn{1}{c|}{}                        & 2               & 4               & 8              & 2                & 4               & 8               \\ \hline
\multirow{2}{*}{ResNet18}   & LSQ\cite{esser2019learned}                                          & 94.9            & 95.2            & 95.3           & 76.4             & 77.3            & 77.2            \\
                            & diffQ\cite{defossez2021differentiable}*                                        & 93.9            & 94.9            & ---              & 71.7             & 77.6            & ---               \\
(FP 95.2/77.7)              & Ours                                         & 95.1            & 95.5            & 95.4           & 76.6             & 77.6            & 77.4            \\ \hline
\multirow{2}{*}{WideResNet} & LSQ\cite{esser2019learned}                                          & 95              & 95.4            & 95.3           & 76               & 77.1            & 77.5            \\
                            & diffQ\cite{defossez2021differentiable}*                                        & 94.1            & 94.6            & ---              & 75.6             & 76.9            & ---               \\
(FP 95.6/77.4)              & Ours                                         & 95.3            & 95.5            & 95.5           & 76.6             & 77.3            & 77.7            \\ \hline
\end{tabular}
\label{tab:cifar}

\end{table}

\begin{table*}[t]
\begin{center}
\caption{Accuracy comparison to the state-of-the-art quantization methods with ResNet structure on ImageNet dataset. Techniques under comparison LSQ~\cite{esser2019learned}, diffQ~\cite{defossez2021differentiable}, QIL~\cite{qil2018}, FAQ~\cite{mckinstry2019discovering}, NICE~\cite{baskin2021nice}, PACT~\cite{choi2018pact}, LQ-Nets~\cite{zhang2018lq}. We also added the state-of-the-art non-uniform quantization method, LCQ~\cite{yamamoto2021learnable}, as a reference. $*$ denotes the result with variable bit-width but with equivalent model size under that precision level.}
\begin{tabular}{c|c|*{6}{w{c}{1.0cm}}}
\hline
          &                 & \multicolumn{3}{c}{Top-1 Acc @ Precision} & \multicolumn{3}{c}{Top-5 Acc @ Precision} \\
Network   & Method          & 2              & 4             & 8             & 2              & 4             & 8             \\ \hline
ResNet-18 &                 & \multicolumn{3}{c}{Full precision: 69.7}       & \multicolumn{3}{c}{Full precision: 89.0}       \\
          & Ours            & \textbf{68.4}  & 70.5          & \textbf{70.8} & \textbf{88.3}  & \textbf{89.6} & \textbf{89.8} \\
          & LSQ (re-impl)           & 68.2           & 70.4          &  70.7         & 88.1           & 89.4          & 89.7          \\
          & diffQ\cite{defossez2021differentiable}*          &      ---          & \textbf{71.1} &     ---         &     ---           &       ---        &    ---           \\
          & QIL\cite{qil2018}             & 65.7           & 70.1          &       ---        &    ---            &   ---            &    ---           \\
          & FAQ\cite{mckinstry2019discovering}             &         ---       & 69.8          & 70.0          &       ---         & 89.1          & 89.3          \\
          & LQ-Nets\cite{zhang2018lq}         & 64.9           & 69.3          &     ---          & 85.9           & 88.8          &     ---          \\
          & PACT\cite{choi2018pact}            & 64.4           & 69.2          &     ---          & 85.6           & 89.0          &       ---        \\
          & Nice\cite{baskin2021nice}            &    ---            & 69.8          &      ---         &  ---              & 89.2          &      ---         \\ \hline
non-uniform ref         & LCQ\cite{yamamoto2021learnable}             & 68.9           & 71.5          &           ---    &   ---             &      ---         &    ---           \\ \hline
ResNet-34 &                 & \multicolumn{3}{c}{Full precision: 75.1}       & \multicolumn{3}{c}{Full precision: 92.3}       \\
          & Ours            & \textbf{72.6}           & \textbf{75.0}          & \textbf{74.9}          & \textbf{90.7}           & \textbf{92.1}          & \textbf{92.3}          \\
          & LSQ (re-impl)             & 72.3           & 74.4          & 74.7          & 90.5           & 92.0          & 92.2          \\
          & QIL\cite{qil2018}             & 70.6           & 73.7          &     ---          &      ---          &  ---             &     ---          \\
          & LQ-Nets\cite{zhang2018lq}         & 69.8           &          ---     &      ---         & 89.1           &          ---     &      ---         \\
          & NICE\cite{baskin2021nice}            &           ---     & 73.5          &     ---          &      ---          & 91.4          &     ---          \\ \hline
non-uniform ref         & LCQ\cite{yamamoto2021learnable}             & 72.7          & 74.3          &    ---           &    ---            &   ---            &     ---          \\ \hline
ResNet-50 &                 & \multicolumn{3}{c}{Full precision: 79.0}       & \multicolumn{3}{c}{Full precision: 94.4}       \\
          & Ours            & \textbf{76.7}  & \textbf{78.4}  & \textbf{78.6} & \textbf{93.2} & \textbf{94.1} & \textbf{94.2}          \\
          & LSQ (re-impl)             & 75.6           & 77.5          & 77.7          & 92.6           & 93.7          & 93.7        \\
          & diffQ\cite{defossez2021differentiable}*          & 76.3           & 76.6          &    ---          &      ---          &  ---             &        ---       \\
          & PACT\cite{choi2018pact}            & 72.2           & 76.5          &     ---          & 90.5           & 93.2          &         ---      \\
          & NICE\cite{baskin2021nice}            &       ---         & 76.5          &        ---       &     ---           & 93.3          &       ---        \\
          & FAQ\cite{mckinstry2019discovering}             &         ---       & 76.3          & 76.5          &      ---          & 92.9          & 93.1          \\ \hline
non-uniform ref         & LCQ\cite{yamamoto2021learnable}            & 75.1           &  76.6        &     ---      &      ---      & ---              &         ---  \\ \hline
\end{tabular}

\label{tab:imagenet}
\end{center}
\end{table*}

Table \ref{tab:cifar} compares the accuracy of the proposed and two other SOTA quantization methods at three different bit-widths for the CIFAR-10 and CIFAR-100 dataset.
As we can see, our method could outperform other SOTA methods. 
Note here we reimplemented LSQ method with the exact setup (same hyper-parameters, same preprocess) as described in their paper, since the original implementation is not available. 

On ImageNet dataset, we compare our method's accuracy with state-of-the-art quantized networks and full precision baselines, as is shown in Table~\ref{tab:imagenet}. 
To facilitate fair comparison, we only consider published works that quantize all the layer weights to the specified precision.
In some cases, we reported higher accuracy than the original publication since our reimplementation of their approach surpassed what they claimed. 
We notice the baseline performance of FP32 models reported by previous quantization models varies about 2\%, e.g. ~\cite{liu2021nonuniformtouniform}'s basline ResNet18 FP model is ($71.8\% > 69.7\%$) 2.1\% better than the \texttt{timm}~\cite{wightman2021resnet} baseline which was using the PyTorch baseline.\footnote{\url{https://pytorch.org/hub/pytorch_vision_resnet/}}
This is significant, since the results of quantized models are usually compared to its FP baseline, missing it by a tiny margin (within 1\%).
Give a fluctuating baseline, it is difficult to gauge how much better a quantization method is.
All of our FP model baseline results are standard numbers which is reported at the \texttt{timm} library readme page.~\footnote{\url{https://rwightman.github.io/pytorch-image-models/results/}}

As is shown in Table~\ref{tab:imagenet}, our proposed method could outperform LSQ across the board.
When the model complexity goes up, and the bits go up, our method could clearly outperform all previous methods.
Our Resnet-18 model seem to get capped off by a relatively weaker basline FP model, consequently, we were not able to reproduce the 71.1\% accurarcy as claimed by LSQ and DiffQ for Resnet-18 4-bit. 
The performance of our reimplemented LSQ only reaches 70.4\%.
However, our 2-bit quantization of Resnet-18 only miss 1.3\% from the full precision model and our 8-bit quantization could surpass the FP model by 1.1\%.
We found that our model achieved a higher top-5 accuracy than all previous reported approaches for 2-, 4- and 8- bit networks with the architectures considered here.
For nearly all cases, our methods achieved top 8-bit to-date performance. 
Interestingly, our method as a uniform quantization method could match the performance of non-uniform quantization method LCQ~\cite{yamamoto2021learnable}. 
\begin{figure}[h]
\centering
\includegraphics[width=0.5\linewidth]{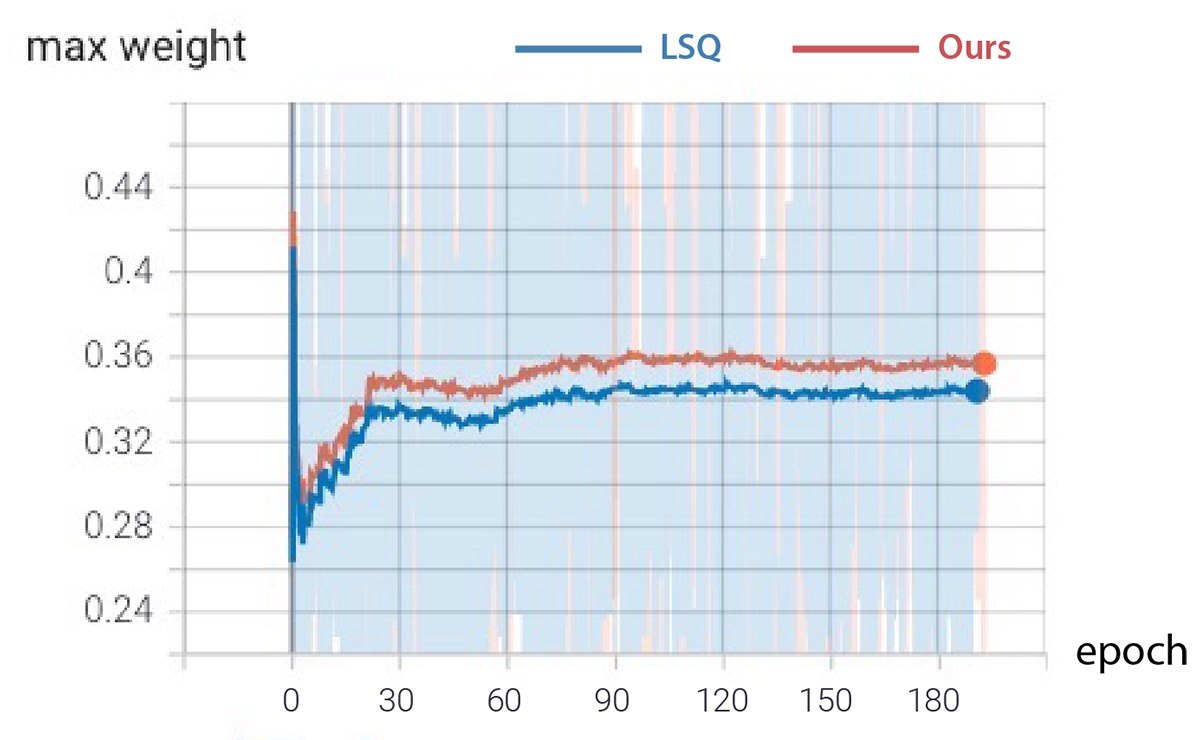}
\caption{Max Weight LSQ Vs Our method. Smoothing 0.999, the background indicates the range of values without smoothing.}
\label{fig:maxweight}
\end{figure}
In practice, the need to run larger models on the edge devices are more of value to the task of quantization.
For instance, running a ResNet-18 model ($<10MB$) is already possible on vast majority the SOTA hardwares without quantization.
In contrast, to shrink a well-performing large-model like ResNet-50 to fit on an edge device should be the core drive behind model quantization.
In the light of this statement, we argue that our method should be of more practical value than all the existing quantization methods.

\section{Discussion}
\subsection{Flatness of Local Optima}
In order to verify that our quantization error aware noise tempering indeed allows our model to converge to a flatter local optima, we picked our best-performing ResNet18 model checkpoints at various precision on CIFAR 100, and measure the local loss landscape sharpness. We adopt the definition of sharpness from \cite{keskar2017large} to compute the maximum value of the loss function in a constrained neighborhood around the minima, to avoid computationally expensive task of computing eigenvalues of $\nabla^2\mathcal{L}(x)$. Specifically, this is achieved through projection onto a $\ell_\infty$ norm ball with radius $\rho$ around the weight parameter $W$: 
$\max_{\|\epsilon \|_\infty < \rho} \mathcal{L}(W + \epsilon) - \mathcal{L}(W)
$\\
\looseness=-1 We follow the implementation from ~\cite{lll2021}
to compute sharpness score of the checkpoints trained with our proposed noise versus without noise. Our results are summarized in Table~\ref{tab:sharpness}. As we can see, the models trained with noise tend to achieve lower sharpness score, indicating a flatter minima, which is favorable for generalization. This also echoes with the better performance of our models reported in Table~\ref{tab:cifar}

\begin{table}[H]
\caption{Sharpness with and without noise tempering (lower is better)}
\centering
\begin{tabular}{l|l|c|*{3}{w{c}{0.9cm}}}
\hline
\multirow{2}{*}{Model}     & \multirow{2}{*}{Noise Tempering} & \multirow{2}{*}{$\rho$} & \multicolumn{3}{c}{sharpness @ Precision $\downarrow$} \\
                           &                                  &                      & 2          & 4             & 8            \\ \hline
\multirow{4}{*}{ResNet 18} & \multirow{2}{*}{w/o tempering}   & 1.00E-03             &   6.09         & 4.08          & 3.85         \\
                           &                                  & 5.00E-04             &  2.78           & \textbf{1.92}          & 1.80          \\ \cline{2-6} 
                           & \multirow{2}{*}{with tempering}  & 1.00E-03             &  \textbf{4.97}         & \textbf{3.93}          & \textbf{3.44}         \\
                           &                                  & 5.00E-04             &  \textbf{2.58}       & \textbf{1.92}          & \textbf{1.65}         \\ \hline
\end{tabular}
\label{tab:sharpness}
\end{table}

\subsection{Effect of hyperparameters}
\begin{table}[H]
\caption{Accuracy at different quantization noise levels.}
\centering
\begin{tabular}{l|l|*{3}{w{c}{0.9cm}}}
\hline
\multirow{2}{*}{Model} & \multicolumn{1}{c|}{\multirow{2}{*}{Noise Level}} & \multicolumn{3}{l}{Top-1 Acc @ Precision} \\
                       & \multicolumn{1}{c|}{}                             &  2              &  4             &  8             \\ \hline
ResNet 18              & 0 (LSQ)                                           & 76.4           & 77.3          & 77.2          \\
(FP 77.7)              & 0.1                                               & 76.6           & 77.3          & \textit{75.3}          \\
                       & 0.2                                               & 76.6           & \textbf{77.6}          & 77.4          \\
                       & 0.3                                               & 76.8           & 77.4          & 77.6          \\
                       & 0.4                                               & \textbf{77.4}           & 77.5          & \textbf{77.7}          \\ \hline
\end{tabular}
\label{tab:performance}
\end{table}
In our proposed method, we introduced two hyperparameters, $c$ and $k$, as in Eq. \ref{eqn:ours}. The hyperparameter $c$ controls the how much the noise temperature is raised at perticular quantization error. The hyperparameter $k$, on the other hand, controls when the system should raise the temperature during the training process. Interestingly, we found that a hyperparameter $k \le 50$ usually does not impact the end performance by too much as long as the model is trained long enough, even when $k=0$, in which case the temperature is raised based on quantization error right at the begining of QAT finetuning. Thus, we set the $k=50$ for all experiments allowing faster convergence and mainly studied the effect of $c$ here.

To investigate how this hyper-parameter influences the accuracy, we attempted different noise level ranging from 0, which is equivalent to LSQ \cite{esser2019learned}, to 0.4 on CIFAR-100 and summarize our result in table 5 and Appendix A.4.
Evidently, in most cases, adding the noise like our method would improve the performance.
Even though there seems to be an optimal noise level for each model-precision pair, it is usually still possible to benefit from sub-optimal noise level.
In our experience, it seems that a noise level factor $c = 0.2 \sim 0.4$ is usually a good option and generate consistently better results on CIFAR 10, CIFAR 100, and ImageNet.

\subsection{Step-size and quantization error and weight}    

Other than the effect on accuracy, we also seek to understand how the noise impact metrics including quantization error and step-size parameter $s$ as described in equation~(\ref{sq}). 
Quantization errors and the scale of weights are shown in Table 6(a,b) in the Appendix. A.5
As is shown, scales and quantization errors are closely correlated.
The complexity of the models do not seem to affect the quantization error, and they are only negatively correlated to the bit-width. 
The more bits, the smaller the scale and thus the smaller quantization error.
Compared to LSQ method, it seems our method tends to slightly inflate the quantization error to balance with the task loss.
Figure~\ref{fig:maxweight} shows that the mean value of the max weights of our methods is larger than LSQ methods across the board.
This indicates that our method might be better at capturing larger weights than LSQ.

\section{Closing Statements}
In summary, our proposal to in-corporate a pseudo quantization noise together with a learnable scale of task loss gradient seems to strike a balance between minimizing the task loss and minimizing the quantization error.
Our experiments demonstrate that our method outperform existing methods with a visible margin. 
Without introducing additional complexity or inducing extra training time, our method seems to bring a ``free lunch" for an extra 1\% accuracy gain. \\
\looseness=-1\textbf{Limitation:} We showed our methods to work with SGD according to clear theoretical motivation. However, for adaptive optimizers such as Adam, AdamW, further analysis is needed. Loss Landscape would also change drastically with adaptive optimizer, and so does all the weight updates during quantization.\\
\textbf{Broader Impact:} Our method is simple to implement and can be readily reproduced in existing deep learning frameworks such as PyTorch, and TensorFlow. 
It is widely applicable to a variety ML models to improve their efficiency, the users tradeoff bit-width and accuracy as their needs.
Furthermore, our theoretical intuition of using noise could potentially inspire more creative methods in engineering noise to suit other tasks.
\newpage

\bibliography{example_paper}

\begin{thebibliography}{10}

\bibitem{baskin2021nice}
Chaim Baskin, Evgenii Zheltonozhkii, Tal Rozen, Natan Liss, Yoav Chai, Eli
  Schwartz, Raja Giryes, Alexander~M Bronstein, and Avi Mendelson.
\newblock Nice: Noise injection and clamping estimation for neural network
  quantization.
\newblock {\em Mathematics}, 9(17):2144, 2021.

\bibitem{bengio2013estimating}
Yoshua Bengio, Nicholas L{\'e}onard, and Aaron Courville.
\newblock Estimating or propagating gradients through stochastic neurons for
  conditional computation.
\newblock {\em arXiv preprint arXiv:1308.3432}, 2013.

\bibitem{bhalgat2020lsq}
Yash Bhalgat, Jinwon Lee, Markus Nagel, Tijmen Blankevoort, and Nojun Kwak.
\newblock Lsq+: Improving low-bit quantization through learnable offsets and
  better initialization, 2020.

\bibitem{randomnessSGD2021}
Arwen~V. Bradley and Carlos~Alberto Gomez-Uribe.
\newblock How can increased randomness in stochastic gradient descent improve
  generalization?, 2021.

\bibitem{cai2020zeroq}
Yaohui Cai, Zhewei Yao, Zhen Dong, Amir Gholami, Michael~W Mahoney, and Kurt
  Keutzer.
\newblock Zeroq: A novel zero shot quantization framework.
\newblock In {\em Proceedings of the IEEE/CVF Conference on Computer Vision and
  Pattern Recognition}, pages 13169--13178, 2020.

\bibitem{pact}
Jungwook Choi, Zhuo Wang, Swagath Venkataramani, Pierce~I{-}Jen Chuang,
  Vijayalakshmi Srinivasan, and Kailash Gopalakrishnan.
\newblock {PACT:} parameterized clipping activation for quantized neural
  networks.
\newblock {\em CoRR}, abs/1805.06085, 2018.

\bibitem{choi2018pact}
Jungwook Choi, Zhuo Wang, Swagath Venkataramani, Pierce I-Jen Chuang,
  Vijayalakshmi Srinivasan, and Kailash Gopalakrishnan.
\newblock Pact: Parameterized clipping activation for quantized neural
  networks.
\newblock {\em arXiv preprint arXiv:1805.06085}, 2018.

\bibitem{courbariaux2015binaryconnect}
Matthieu Courbariaux, Yoshua Bengio, and Jean-Pierre David.
\newblock Binaryconnect: Training deep neural networks with binary weights
  during propagations.
\newblock In {\em Advances in neural information processing systems}, pages
  3123--3131, 2015.

\bibitem{defossez2021differentiable}
Alexandre D{\'e}fossez, Yossi Adi, and Gabriel Synnaeve.
\newblock Differentiable model compression via pseudo quantization noise.
\newblock {\em arXiv preprint arXiv:2104.09987}, 2021.

\bibitem{esser2019learned}
Steven~K Esser, Jeffrey~L McKinstry, Deepika Bablani, Rathinakumar Appuswamy,
  and Dharmendra~S Modha.
\newblock Learned step size quantization.
\newblock {\em arXiv preprint arXiv:1902.08153}, 2019.

\bibitem{fan2021training}
Angela Fan, Pierre Stock, Benjamin Graham, Edouard Grave, Remi Gribonval, Herve
  Jegou, and Armand Joulin.
\newblock Training with quantization noise for extreme model compression, 2021.

\bibitem{gholami2021survey}
Amir Gholami, Sehoon Kim, Zhen Dong, Zhewei Yao, Michael~W Mahoney, and Kurt
  Keutzer.
\newblock A survey of quantization methods for efficient neural network
  inference.
\newblock {\em arXiv preprint arXiv:2103.13630}, 2021.

\bibitem{he2019rethinking}
Kaiming He, Ross Girshick, and Piotr Doll{\'a}r.
\newblock Rethinking imagenet pre-training.
\newblock In {\em Proceedings of the IEEE/CVF International Conference on
  Computer Vision}, pages 4918--4927, 2019.

\bibitem{jacob2018quantization}
Benoit Jacob, Skirmantas Kligys, Bo~Chen, Menglong Zhu, Matthew Tang, Andrew
  Howard, Hartwig Adam, and Dmitry Kalenichenko.
\newblock Quantization and training of neural networks for efficient
  integer-arithmetic-only inference.
\newblock In {\em Proceedings of the IEEE Conference on Computer Vision and
  Pattern Recognition}, pages 2704--2713, 2018.

\bibitem{jain2019trained}
Sambhav Jain, Albert Gural, Michael Wu, and Chris Dick.
\newblock Trained quantization thresholds for accurate and efficient
  fixed-point inference of deep neural networks.
\newblock In I.~Dhillon, D.~Papailiopoulos, and V.~Sze, editors, {\em
  Proceedings of Machine Learning and Systems}, volume~2, pages 112--128. 2020.

\bibitem{qil2018}
Sangil Jung, Changyong Son, Seohyung Lee, JinWoo Son, Youngjun Kwak, Jae{-}Joon
  Han, and Changkyu Choi.
\newblock Joint training of low-precision neural network with quantization
  interval parameters.
\newblock {\em CoRR}, abs/1808.05779, 2018.

\bibitem{keskar2017large}
Nitish~Shirish Keskar, Dheevatsa Mudigere, Jorge Nocedal, Mikhail Smelyanskiy,
  and Ping Tak~Peter Tang.
\newblock On large-batch training for deep learning: Generalization gap and
  sharp minima.
\newblock In {\em International Conference on Learning Representations}, 2017.

\bibitem{li2021brecq}
Yuhang Li, Ruihao Gong, Xu~Tan, Yang Yang, Peng Hu, Qi~Zhang, Fengwei Yu, Wei
  Wang, and Shi Gu.
\newblock Brecq: Pushing the limit of post-training quantization by block
  reconstruction.
\newblock 2021.

\bibitem{liu2021nonuniformtouniform}
Zechun Liu, Kwang-Ting Cheng, Dong Huang, Eric Xing, and Zhiqiang Shen.
\newblock Nonuniform-to-uniform quantization: Towards accurate quantization via
  generalized straight-through estimation.
\newblock 2021.

\bibitem{mckinstry2019discovering}
Jeffrey~L. McKinstry, Steven~K. Esser, Rathinakumar Appuswamy, Deepika Bablani,
  John~V. Arthur, Izzet~B. Yildiz, and Dharmendra~S. Modha.
\newblock Discovering low-precision networks close to full-precision networks
  for efficient embedded inference, 2019.

\bibitem{lll2021}
Sanket~Vaibhav Mehta, Darshan Patil, Sarath Chandar, and Emma Strubell.
\newblock An empirical investigation of the role of pre-training in lifelong
  learning.
\newblock {\em CoRR}, abs/2112.09153, 2021.

\bibitem{nagel2020up}
Markus Nagel, Rana~Ali Amjad, Mart Van~Baalen, Christos Louizos, and Tijmen
  Blankevoort.
\newblock Up or down? adaptive rounding for post-training quantization.
\newblock In {\em International Conference on Machine Learning}, pages
  7197--7206. PMLR, 2020.

\bibitem{nahshan2020loss}
Yury Nahshan, Brian Chmiel, Chaim Baskin, Evgenii Zheltonozhskii, Ron Banner,
  Alex~M. Bronstein, and Avi Mendelson.
\newblock Loss aware post-training quantization, 2020.

\bibitem{neill2020overview}
James~O' Neill.
\newblock An overview of neural network compression.
\newblock {\em arXiv preprint arXiv:2006.03669}, 2020.

\bibitem{sakr2018per}
Charbel Sakr and Naresh Shanbhag.
\newblock Per-tensor fixed-point quantization of the back-propagation
  algorithm.
\newblock {\em arXiv preprint arXiv:1812.11732}, 2018.

\bibitem{smith1983comparison}
M~Smith and Y~Guo.
\newblock A comparison of methods for randomizing phase quantization errors in
  phased arrays.
\newblock {\em IEEE transactions on antennas and propagation}, 31(6):821--828,
  1983.

\bibitem{regSGD}
Samuel~L. Smith, Benoit Dherin, David G.~T. Barrett, and Soham De.
\newblock On the origin of implicit regularization in stochastic gradient
  descent.
\newblock {\em CoRR}, abs/2101.12176, 2021.

\bibitem{noiseSGD2020}
Samuel~L. Smith, Erich Elsen, and Soham De.
\newblock On the generalization benefit of noise in stochastic gradient
  descent.
\newblock {\em CoRR}, abs/2006.15081, 2020.

\bibitem{dontdecaylr2017}
Samuel~L. Smith, Pieter{-}Jan Kindermans, and Quoc~V. Le.
\newblock Don't decay the learning rate, increase the batch size.
\newblock {\em CoRR}, abs/1711.00489, 2017.

\bibitem{bayesSGD2017}
Samuel~L. Smith and Quoc~V. Le.
\newblock A bayesian perspective on generalization and stochastic gradient
  descent.
\newblock {\em CoRR}, abs/1710.06451, 2017.

\bibitem{stock2020and}
Pierre Stock, Armand Joulin, R{\'e}mi Gribonval, Benjamin Graham, and Herv{\'e}
  J{\'e}gou.
\newblock And the bit goes down: Revisiting the quantization of neural
  networks.
\newblock In {\em Eighth International Conference on Learning Representations},
  2020.

\bibitem{taheri2008}
S.~Taheri and F.~Farzaneh.
\newblock New methods of reducing the phase quantization error effects on beam
  pointing and parasitic side lobe level of the phased array antennas.
\newblock In {\em 2006 Asia-Pacific Microwave Conference}, pages 2114--2117,
  2006.

\bibitem{uhlich2019mixed}
Stefan Uhlich, Lukas Mauch, Fabien Cardinaux, Kazuki Yoshiyama, Javier~Alonso
  Garcia, Stephen Tiedemann, Thomas Kemp, and Akira Nakamura.
\newblock Mixed precision dnns: All you need is a good parametrization.
\newblock In {\em International Conference on Learning Representations}, 2020.

\bibitem{wightman2021resnet}
Ross Wightman, Hugo Touvron, and Herv{\'e} J{\'e}gou.
\newblock Resnet strikes back: An improved training procedure in timm.
\newblock {\em arXiv preprint arXiv:2110.00476}, 2021.

\bibitem{yamamoto2021learnable}
Kohei Yamamoto.
\newblock Learnable companding quantization for accurate low-bit neural
  networks.
\newblock In {\em Proceedings of the IEEE/CVF Conference on Computer Vision and
  Pattern Recognition}, pages 5029--5038, 2021.

\bibitem{yin2019understanding}
Penghang Yin, Jiancheng Lyu, Shuai Zhang, Stanley Osher, Yingyong Qi, and Jack
  Xin.
\newblock Understanding straight-through estimator in training activation
  quantized neural nets, 2019.

\bibitem{zhang2018lq}
Dongqing Zhang, Jiaolong Yang, Dongqiangzi Ye, and Gang Hua.
\newblock Lq-nets: Learned quantization for highly accurate and compact deep
  neural networks.
\newblock In {\em Proceedings of the European conference on computer vision
  (ECCV)}, pages 365--382, 2018.

\bibitem{zhou2016dorefa}
Shuchang Zhou, Yuxin Wu, Zekun Ni, Xinyu Zhou, He~Wen, and Yuheng Zou.
\newblock Dorefa-net: Training low bitwidth convolutional neural networks with
  low bitwidth gradients.
\newblock {\em arXiv preprint arXiv:1606.06160}, 2016.

\end{thebibliography}
\bibliographystyle{plain}

\appendix

\newpage
\section{Appendix}

\begin{table}[t]
\centering
\renewcommand{\arraystretch}{1.5}
\begin{tabular}{|c|c|c|}
\hline
\textbf{Symbol} & \textbf{Meaning} & \textbf{Type}\\ \hline
$N$      & total number of samples of training data & scalar  \\ \hline
$B$      & miniBatch Size & scalar  \\ \hline
$m$      & momentum coefficient   & scalar   \\ \hline
$W_t$      & weights at step $t$   & vector    \\ \hline
$\varepsilon_\Phi$, $\tilde{\varepsilon}_\Phi$      & independent isotropic Gaussian noise   & vector    \\ \hline
$\varepsilon(t)$      & mini-batch gradient noise at step $t$  & vector    \\ \hline
$\displaystyle \tilde{\mathbf{Q}}(W)$      & quantized weights with noise   & vector    \\ \hline
$\displaystyle \mathbf{Q}(W) $     & quantized weights  & vector    \\ \hline
$\displaystyle \nabla_{\tilde{\mathbf{Q}}(W)}\hat{\mathcal{L}}$      & mini-batch gradient w.r.t. the quantized value with noise & vector     \\ \hline
$\displaystyle \nabla_{\mathbf{Q}(W)}\hat{\mathcal{L}}$      & mini-batch gradient w.r.t. the quantized value \emph{without} noise  & vector    \\ \hline
$\displaystyle \nabla_{\tilde{\mathbf{Q}}(W)}\mathcal{L}$      & full-batch gradient w.r.t. the quantized value with noise  & vector    \\ \hline
$\displaystyle \nabla_{\mathbf{Q}(W)}\mathcal{L}$      & full-batch gradient w.r.t. the quantized value \emph{without} noise  & vector     \\ \hline
$\displaystyle \nabla^2_{\mathbf{Q}(W)}\mathcal{L}$      & full-batch hessian w.r.t. the quantized value \emph{without} noise & matrix     \\ \hline
\end{tabular}
\renewcommand{\arraystretch}{1}
\end{table}

\subsection{Noise in SGD with momentum}
For SGD with momentum, we can define the ``True Accumulation", $V$, and ``Estimated Accumulation", $\hat{V}$ just like the vanilla SGD. The ``true accumulation" at step $T$ is defined as the the discounted sum of all true gradient from step 0 to step $T$: 
\begin{equation}
    V_T = \sum_{t=0}^{T} m^t \cdot\nabla_{\mathbf{Q}(W_t)}\mathcal{L}
\end{equation}
Likewise, the ``estimated accumulation" is defined as the discounted sum of all estimated gradient from mini-batches:
\begin{equation}
    \hat{V}_T = \sum_{t=0}^{T} m^t \cdot\nabla_{\mathbf{Q}(W_t)}\hat{\mathcal{L}} = \sum_{t=0}^{T} m^t \cdot \left(\nabla_{\mathbf{Q}(W_t)}\mathcal{L} + \varepsilon(t)\right)
\end{equation}
Thus, using SGD with momentum, we end up with the following update formula.
\begin{equation}
\begin{aligned}
    \hat{V}_{t+1} &= m\hat{V}_{t} + \nabla_{\mathbf{Q}(W_t)} \hat{\mathcal{L}} \\
    W_{t+1}' &= W_{t}' - \eta \hat{V}_{t+1}
\end{aligned}
\label{sgdm}
\end{equation}
We can expand Eq. \ref{sgdm} using telescoping series and obtain the estimated accumulation using true accumulation and a combination of Gaussian noise samples.
\begin{equation}
    \hat{V}_T = \sum_{t=0}^T m^t \cdot \nabla_{\mathbf{Q}(W_t)}\mathcal{L} + \sum_{t=0}^T \varepsilon(t) \cdot m^t = V_T + \sum_{t=0}^T \varepsilon(t) \cdot m^t
\end{equation}
Assuming each batches are drawn independently at random (in reality, this approximation become more accurate when $N \gg B$ and $N \rightarrow \infty$). Under this assumption and suppose we are using gradient tempering in Eq. 4 (here, we let the noise $\varepsilon(t)$ to automatically include the tempering term we manually added. In general, and following derivations, this will represent the mini-batch noise \emph{without} tempering), the term $\sum_{t=0}^T \varepsilon(t) \cdot m^t$ is then an independent sum of Gaussian noises, each has mean 0 and variance of $\Sigma(W_t')/B + \gamma_{Q_t}^2 T_{Q_t}$. Thus, if we assume the variance of $\varepsilon(t)$ at each step does not change too much, the decaying sum of noise term has total variance of:
\begin{equation}
\begin{aligned}
    \sum_{t=0}^T m^{2t}\left(\Sigma(W_t')/B + \gamma_{Q_t}^2 T_{Q_t}\right) &\le \sum_{t=0}^\infty m^{2t}\left( \Sigma(W_t')/B + \gamma_{Q_t}^2 T_{Q_t} \right)\\
    &\approx \left( \Sigma(W_T')/B + \gamma_{Q_T}^2 T_{Q_T} \right) \sum_{t=0}^T m^{2t}\\
    &= \frac{1}{1-m^2}\left( \Sigma(W_T')/B + \gamma_{Q_T}^2 T_{Q_T} \right) 
\end{aligned}
\end{equation}

In this cases, the momentum in SGD is effectively serving as a \emph{scaling factor} depending on the momentum coefficient $m$. It simultaneously scales both the gradient noise due to mini-batch sampling and quantization-error dependent noise which we introduced.

\subsection{Relating gradient tempering to quantization function}

To show that our choice $\gamma_Q$ and $T_Q$ in \S 3.2 results in the quantization function as in Eq. 7, we assume that the hyperparameters $k$ and $c$ are choosen so that when tempering is active, the quantization error is relatively small (due to step-size parameter $s$ converging to a relatively small value). We further assume that the gradient noise at step $t$, $\varepsilon(t)$, has similar distribution at $\mathbf{Q}(W_t)$ and $\tilde{\mathbf{Q}}(W_t)$. This allows us to use first order Taylor expansion to arrive at the following approximation:
\begin{equation}
    \begin{aligned}
        \nabla_{\tilde{\mathbf{Q}}(W_t)}\hat{\mathcal{L}} &= \nabla_{\tilde{\mathbf{Q}}(W_t)}\mathcal{L} + \varepsilon(t)\\
        &\approx \left[ \nabla_{\mathbf{Q}(W_t)}\mathcal{L} + \nabla^2_{\mathbf{Q}(W_t)}\mathcal{L} \cdot \left(\tilde{\mathbf{Q}}(W_t) - \mathbf{Q} (W_t)\right) \right] + \varepsilon(t)\\
        &= \left[ \nabla_{\mathbf{Q}(W_t)}\mathcal{L} +  \nabla^2_{\mathbf{Q}(W_t)}\mathcal{L} \cdot \left(c\exp(-k|\mathbf{Q}(W_t) - W_t |) \sqrt{|\mathbf{Q}(W_t) - W_t |} \tilde{\varepsilon}_\Phi \right) \right] + \varepsilon(t)
    \end{aligned}
\end{equation}
Here the term $c\exp(-k|\mathbf{Q}(W) - W |)$ and $\sqrt{|\mathbf{Q}(W) - W |}$ are diagonal matrixes that control the variance of the noise. Thus, we break down the terms further into $\gamma_Q$ and $T_Q$ with the following:
\begin{equation}
    \begin{aligned}
        \nabla_{\tilde{\mathbf{Q}}(W_t)}\hat{\mathcal{L}} &\approx \left[ \nabla_{\mathbf{Q}(W_t)}\mathcal{L} +  \nabla^2_{\mathbf{Q}(W_t)}\mathcal{L} \cdot \left(c\exp(-k|\mathbf{Q}(W_t) - W_t |) \sqrt{|\mathbf{Q}(W_t) - W_t |} \tilde{\varepsilon}_\Phi \right) \right] + \varepsilon(t)\\
        &= \nabla_{\mathbf{Q}(W_t)}\mathcal{L} +  \underbrace{c\exp(-k|\mathbf{Q}(W_t) - W_t |) \nabla^2_{\mathbf{Q}(W_t)}\mathcal{L} }_{\gamma_Q} \cdot \underbrace{\sqrt{|\mathbf{Q}(W_t) - W_t |}}_{\sqrt{T_Q}} \tilde{\varepsilon}_\Phi + \varepsilon(t)\\
        &=\nabla_{\mathbf{Q}(W_t)}\mathcal{L} + \gamma_Q \sqrt{T_Q} \tilde{\varepsilon}_\Phi + \varepsilon(t)
    \end{aligned}
\end{equation}
Taking $\eta$ as learning rate, we can then arrive at the update as in Eq. 4 using assumption that mini-batch gradient noise $\varepsilon(t)$ is approximately same at $\mathbf{Q}(W)$ and $\tilde{\mathbf{Q}}(W)$
\begin{equation}
    \begin{aligned}
        W_{t+1} &= W_t - \eta\nabla_{\tilde{\mathbf{Q}}(W_t)}\hat{\mathcal{L}}\\
        &\approx W_t -\eta\nabla_{\mathbf{Q}(W_t)}\mathcal{L} + \eta\gamma_Q \sqrt{T_Q} \tilde{\varepsilon}_\Phi + \eta\varepsilon(t)\\
        &=  \underbrace{W_t -\eta\nabla_{\mathbf{Q}(W_t)}\mathcal{L} + \eta\gamma_Q \sqrt{T_Q} \tilde{\varepsilon}_\Phi + \sqrt{\eta T}\Sigma(W_t)^{1/2}\varepsilon_\Phi}_{\text{Update in Equation 4}}
    \end{aligned}
\end{equation}
Notice that this proof uses the first order Taylor expansion to perform estimation. However, this estimation can be poor if the noise is large. i.e. violation of local linearity. 
To make sure this is unlikely to happen in our model, our choice of noise term naturally bounds the variance of the noise by the step-size parameter, $s$. 
Since the step-size tends to be small when deeper into the training phase as shown in Table \ref{tab:quantscale}, linear approximation is assured to perform well.

\subsection{Experiment Settings}
\paragraph{Evaluation on CIFAR10/100}
We performed experiments using ResNet-18 and WideResNet model on the CIFAR-10/100 dataset for the ease of comparison to LSQ and DiffQ.
We trained the quantized models over 100 epoches with an initial learning rate of 0.01 for the weights. 
The weight decay was set to 1E-4. 
We adopted standard data augmentation techniques, namely random crop and horizontal flip.

\paragraph{Evaluation on ImageNet}
As for our experiement on ImageNet dataset, images were resized to 256 x 256, then a 224 x 224 crop was selected for training, with horizontal mirroring applied half the time.
At test time, a 224 x 224 centered crop was chosen.
We implemented and tested all of our methods in PyTorch.

In order to have a fair comparison with previous works, we set weights to either 2-, 4-, or 8-bit for all matrix multiplication layers.
All quantized networks are initialized using weights from a pre-trained full precision model load from the \texttt{timm} library~\cite{wightman2021resnet} with equivalent architecture before fine-tuning in the quantized space.

Networks were trained with SGD optimizers, a momentum of 0.9, using a softmax cross entropy loss function, and cosine learning rate annealing without restarts. 
All the networks are trained for 200 epoches, initial learning rate was set to 0.1 for full precision networks, 0.01 for 2-, and 4-bit, 8-bit networks.
On top of adopting Cosine Annealing scheduling, we also reduce learning rate 10 folds after every 50 epoches as fine tuning, which was mentioned by \cite{he2019rethinking} as the best practice.

\subsection{Quantization noise level}
In this section, we provide additional experiment in studying the hyperparameter $c$ in Eq. 7. Results in this table is similar to those in \S 5.2. The best performing $c$ is ranges from $0.2 \sim 0.4$.
\begin{table}[H]
\caption{Accuracy at different quantization noise levels.}
\centering
\begin{tabular}{l|l|*{3}{w{c}{0.9cm}}}
\multirow{2}{*}{Model} & \multicolumn{1}{c|}{\multirow{2}{*}{Noise Level}} & \multicolumn{3}{l}{Top-1 Acc @ Precision} \\
                      & \multicolumn{1}{c|}{}                             &  2              &  4             &  8             \\ \hline
ResNet 50              & 0 (LSQ)                                           & 77.6           & 78.5          & 78.5          \\
(FP 79.2)              & 0.1                                               & \textbf{77.9}           & 78.4          & \textit{78.1}          \\
                      & 0.2                                               & 77.7           & \textbf{78.8}          & \textbf{78.7}          \\
                      & 0.3                                               & 77.8           & 78.4          & 78.4          \\
                      & 0.4                                               & 77.8           & 78.5          & 78.4          \\ \hline
ResNet 152             & 0 (LSQ)                                           & 79.0           & 79.2          & 79.4          \\
(FP 79.9)              & 0.1                                               & 79.0           & 79.6          & 79.5          \\
                      & 0.2                                               & \textbf{79.3}           & \textbf{79.9}          & \textbf{79.6}          \\
                      & 0.3                                               & 79.3           & 79.4          & \textbf{79.6}          \\
                      & 0.4                                               & 79.2           & 79.5          & \textbf{79.6}         
\end{tabular}
\label{tab:noise}
\end{table}

\subsection{Quantization error and step-size parameter at termination}
We also studied the quantization error and step-size parameter after using tempering introduced in Eq. 7, summarized in table \ref{tab:quanterror} and table \ref{tab:quantscale}
\begin{table}[]
\centering
\begin{subtable}[]{0.7\textwidth}
\centering
\caption{Quantization Errors}
\begin{tabular}{c|c|ccc}
          &        & \multicolumn{3}{c}{BitWidth}   \\
Network   & Method & 2        & 4        & 8        \\ \hline
ResNet-18 & w/o tempering    & 9.00E-03         & 5.30E-03         & 6.70E-04         \\
          & Ours   & 9.00E-03 & 5.30E-03 & 6.50E-04 \\ \hline
ResNet-34 & w/o tempering    &  0.011        & 4.00E-03         &   5.0E-04       \\
          & Ours   &  0.012        & 4.40E-03         &   6.0E-04\\ \hline
ResNet-50 & w/o tempering    &  0.011        &  4.10E-03        &   6.0E-04       \\
          & Ours   &  0.011   &   4.9E-03       &   6.5E-04      
\end{tabular}
\label{tab:quanterror}
\end{subtable}
\begin{subtable}[]{0.7\textwidth}
\centering
\caption{step-size parameter}
\begin{tabular}{c|c|ccc}
          &        & \multicolumn{3}{c}{BitWidth}   \\
Network   & Method & 2        & 4        & 8        \\ \hline
ResNet-18 & w/o tempering    &  0.034        &  0.019        &    2.5E-03      \\
          & Ours   & 0.04 & 0.015 & 2.6E-03 \\ \hline
ResNet-34 & w/o tempering    &  0.036        &   0.013       &    2.2E-03      \\
          & Ours   &  0.04        &   0.015       &     2.3E-03     \\ \hline
ResNet-50 & w/o tempering    &   0.035       &   0.015       &  2.4E-03        \\
          & Ours   &   0.04       &   0.017       &  2.6E-03       
\end{tabular}
\label{tab:quantscale}
\end{subtable}
\caption{Comparison between on Quantization errors and step-size parameter with or without tempering}
\end{table}

There seems to be a slight increase in both the step-size and the quantization error after noise tempering. While this might seems to be worrisome, it is \emph{not} necessarily true that large quantization error directly leads to worse performance in the prediction task.

Most uniform quantization methods in previous works, such as diffQ~\citep[pg. 3]{defossez2021differentiable} and LSQ~\cite{esser2019learned}, generally have the element-wise quantization function $\mathbf{Q}(w)$ with the following form, where $\lfloor \cdot \rceil$ represents the $\text{round}(\cdot)$ function:
$$\mathbf{Q}(w) = (\lfloor clip(w/s + b)\rceil -b)\cdot s$$
The parameter $w,\ s,\ b$ may be learned or manually set, but once the training is completed, they are all merely fixed numbers. In our case, we set $b = 0$ and leave $s$ and $w$ as trainable parameters.

One simple observation is that $\mathbf{Q}(w) = \mathbf{Q}(\mathbf{Q}(w))$. The following is the proof when clamping is not happening; the same conclusion still holds when clamping is playing a role:
\begin{equation}
\begin{aligned}
    \mathbf{Q}(\mathbf{Q}(w)) &= (\lfloor\mathbf{Q}(w)/ s + b\rceil-b)\cdot s\\
    &=(\lfloor(\lfloor w/s + b\rceil-b)\cdot s/s + b\rceil-b)\cdot s\\
    &= (\lfloor \lfloor w/s + b \rceil -b + b \rceil -b)\cdot s\\
    &= (\lfloor \lfloor w/s + b \rceil \rceil -b)\cdot s\\
    &= \mathbf{Q}(w)
\end{aligned}    
\end{equation}
Hence, suppose that $w^*$ is optimal of the \emph{quantized} model under loss function $\mathcal{L}$, $\mathbf{Q}(w^*)$ is also an optimal under the same loss function because:
\begin{equation}
\begin{aligned}
    &\quad \mathcal{L}(\mathbf{Q}(\mathbf{Q}(w^*)) + |\mathbf{Q}(\mathbf{Q}(w^*)) - \mathbf{Q}(w^*)|\cdot n,X)\\
    &= \mathcal{L}(\mathbf{Q}(w^*) + 0\cdot n,X)\\
    &= \mathcal{L}(\mathbf{Q}(w^*),X)
\end{aligned}
\label{eqn:opt}
\end{equation}

Using the fact that multiple $w$ will be rounded to $\mathbf{Q}(w)$, as in Fig 2(B), it should be obvious that any $w$ such that $\mathbf{Q}(w) = \mathbf{Q}(w^*)$ are also optimal for the quantized model, albeit with non-zero quantization error.

The only concern raised for larger quantization error is the gradient after STE can be less accurate, which could lead to the model getting trapped at a bad sub-optima. We showed in \S 5.1 that models trained with noise tempering tend to converge to flatter minima.
Converging to a flatter minima, errors in STE gradient would affect the model less compared to models trained without noise tempering (as explained in \S 3.1, stuck at sub-optimal values in a flat basin reduces the loss incurred). 
Empirically, we observe models that trained with noise tempering are less likely to lower the step-size. Instead, the model attempts to expand weight ranges.

\subsection{Future Direction}
Our experiments demonstrate that our method outperform existing methods with a visible margin. 
Without introducing additional complexity or inducing extra training time, our method seems to bring a ``free lunch" for the extra accuracy gain. 
We hope to extend this quantization model to other tasks such as the language and audio tasks, and generalize the performance gain in ResNet to broader range of architectures like transformers.
Going forward, we believe it would be beneficial for the research community to adopt the same baselines using standard libraries like the \texttt{timm} package for fair evaluation as is advocated by~\cite{wightman2021resnet}.

\end{document}